\documentclass{optica-article}
\usepackage{enumitem}
\usepackage[labelfont=bf]{caption}
\usepackage{relsize}
\usepackage{comment}
\usepackage{amsmath}
\usepackage{ulem}
\usepackage{float}
\usepackage{placeins}
\usepackage{tabularray}
\journal{opticajournal} 

\articletype{Research Article}

\usepackage{lineno}
\linenumbers 

\begin{document}
\nolinenumbers
\title{HyperColorization: Propagating spatially sparse noisy spectral clues for reconstructing hyperspectral images}

\author{M. Kerem Aydin,\authormark{1} Qi Guo,\authormark{2} and Emma Alexander\authormark{1,*}}

\address{\authormark{1}Department of Computer Science, McCormick School of Engineering and Applied Science, Northwestern University, 2233 Tech Drive, Evanston, IL, 60208, USA\\
\authormark{2}Elmore Family School of Electrical and Computer Engineering, Purdue University, 501 Northwestern Avenue, West Lafayette, IN, 47907, USA\\
}
\email{\authormark{*}ealexander@northwestern.edu} 


\begin{abstract*} 
Hyperspectral cameras face challenging spatial-spectral resolution trade-offs and are more affected by shot noise than RGB photos taken over the same total exposure time. Here, we present a colorization algorithm to reconstruct hyperspectral images from a grayscale guide image and spatially sparse spectral clues. We demonstrate that our algorithm generalizes to varying spectral dimensions for hyperspectral images, and show that colorizing in a low-rank space reduces compute time and the impact of shot noise. To enhance robustness, we incorporate guided sampling, edge-aware filtering, and dimensionality estimation techniques. Our method surpasses previous algorithms in various performance metrics, including SSIM, PSNR, GFC, and EMD, which we analyze as metrics for characterizing hyperspectral image quality. Collectively, these findings provide a promising avenue for overcoming the time-space-wavelength resolution trade-off by reconstructing a dense hyperspectral image from samples obtained by whisk or push broom scanners, as well as hybrid spatial-spectral computational imaging systems. 
\end{abstract*}

\section{Introduction}
Hyperspectral cameras (HSC) have the potential to significantly improve performance in tasks such as object detection or classification by capturing images with a broader spectral range and higher spectral resolution \cite{yan2021object, li2019deep}. {Example applications include food quality control \cite{feng2012application}, detecting crop disease in agriculture \cite{lu2020recent}, pigment mapping in art inspection \cite{cucci2019hyperspectral}, mineral classification in geology \cite{peyghambari2021hyperspectral}, astronomy \cite{merenyi2014hyperspectral} and tumor detection in biomedical engineering \cite{panasyuk2007medical}. Many of these tasks call for tightly targeted interventions that require high spatial resolution, but in traditional HSCs there is a rigid trade-off between spatial and spectral resolution that cannot adjust to scene content. Additionally, spectral resolution divides light throughput across numerous spectral channels so that HSCs are more susceptible to shot noise compared to similar RGB cameras when exposure times are equal. These drawbacks may explain why HSCs have not seen widespread adoption despite their advantages in many important applications.}



{Traditional hyperspectral cameras acquire three-dimensional data cubes by sequentially capturing two-dimensional slices. In the case of pushbroom cameras, the capture occurs one spatial row at a time, employing a narrow slit and dispersing prism to cover all wavelengths simultaneously \cite{hartley1994linear}. An alternative method involves capturing the entire scene, wavelength by wavelength. While conventional systems typically rely on numerous narrowband filters \cite{wang2014multispectral}, recent advancements, as demonstrated by Zhang et al. \cite{zhang2021deeply}, and Feng et al. \cite{feng2023superposition}, reveal the feasibility of achieving high spectral resolution with a reduced number of broadband spectral filters in concert with an artificial intelligence-driven reconstruction algorithm. However, these approaches still require a sacrifice in spatial or temporal resolution to capture the suggested channels, and may require retraining when used on new datasets.}

{Recent advances in hyperspectral imaging have introduced novel approaches, including compressed sensing and hybrid camera systems. Compressed sensing techniques involve reconstructing hyperspectral images (HSI) from fewer measurements than the actual reconstruction dimensionality. An example is the coded aperture snapshot spectral imager (CASSI), which utilizes a single measurement acquired through a coded aperture and one or two dispersing prisms \cite{wagadarikar2008single,gehm2007single}. Compressed-sensing-based HSIs often exhibit lower spatial resolution and entail solving intricate inverse problems. To address this limitation, hybrid camera systems have emerged, incorporating a second camera dedicated to capturing high-resolution RGB or grayscale images. The introduction of this supporting camera simplifies the inverse problem, as spatial details are handled by the secondary camera. However, the integration of information from both systems presents challenges, requiring careful consideration in the fusion process\cite{saragadam2021sassi, cao2011high, xie2023dual}.}

{Colorization is the process of adding color to black-and-white or grayscale images to create a color version. It is valuable in various fields, including photography, film restoration, and computer graphics. The primary objective of RGB colorization is to augment visual appeal and deliver a more realistic or artistically enhanced portrayal of the original scene. Early endeavors in colorization were manual and labor-intensive, involving handcrafted applications \cite{bonnard2016melies}. In 1983, “Colorization Inc.” pioneered a computer-assisted automated process, marking a turning point in the commercial success of colorization \cite{markle1989method, markle1988coloring, markle1987method}. Over time, propelled by the increasing computational prowess of computers, digital colorization techniques evolved, requiring less user intervention and yielding improved reconstructions. Noteworthy advancements include contributions by Levin et al. \cite{levin2004colorization}, Ironi et al. \cite{ironi2005colorization}, and Yatziv et al. \cite{yatziv2006fast}. The landscape of colorization underwent transformative changes with the introduction of deep learning. Cheng et al. developed a convolutional neural network capable of autonomously colorizing images without human intervention \cite{Cheng_2015_ICCV}. Subsequently, the advent of semantic segmentation enabled pixel-wise classification of objects, paving the way for innovative colorization techniques \cite{larsson2016learning}. Presently, state-of-the-art methods incorporate colorization-specific network architectures \cite{iizuka2016let}, as well as emerging techniques like transformers and attention mechanisms \cite{kumar2021colorization, wan2022bringing}. Even though state-of-the-art AI-based approaches offer higher performance where plenty of data is available, classical algorithms still attract attention when computational resources are limited, interpretability is desired or amount of available data is limited.}

\begin{figure}[ht!]
    \centering\includegraphics[width=10cm]{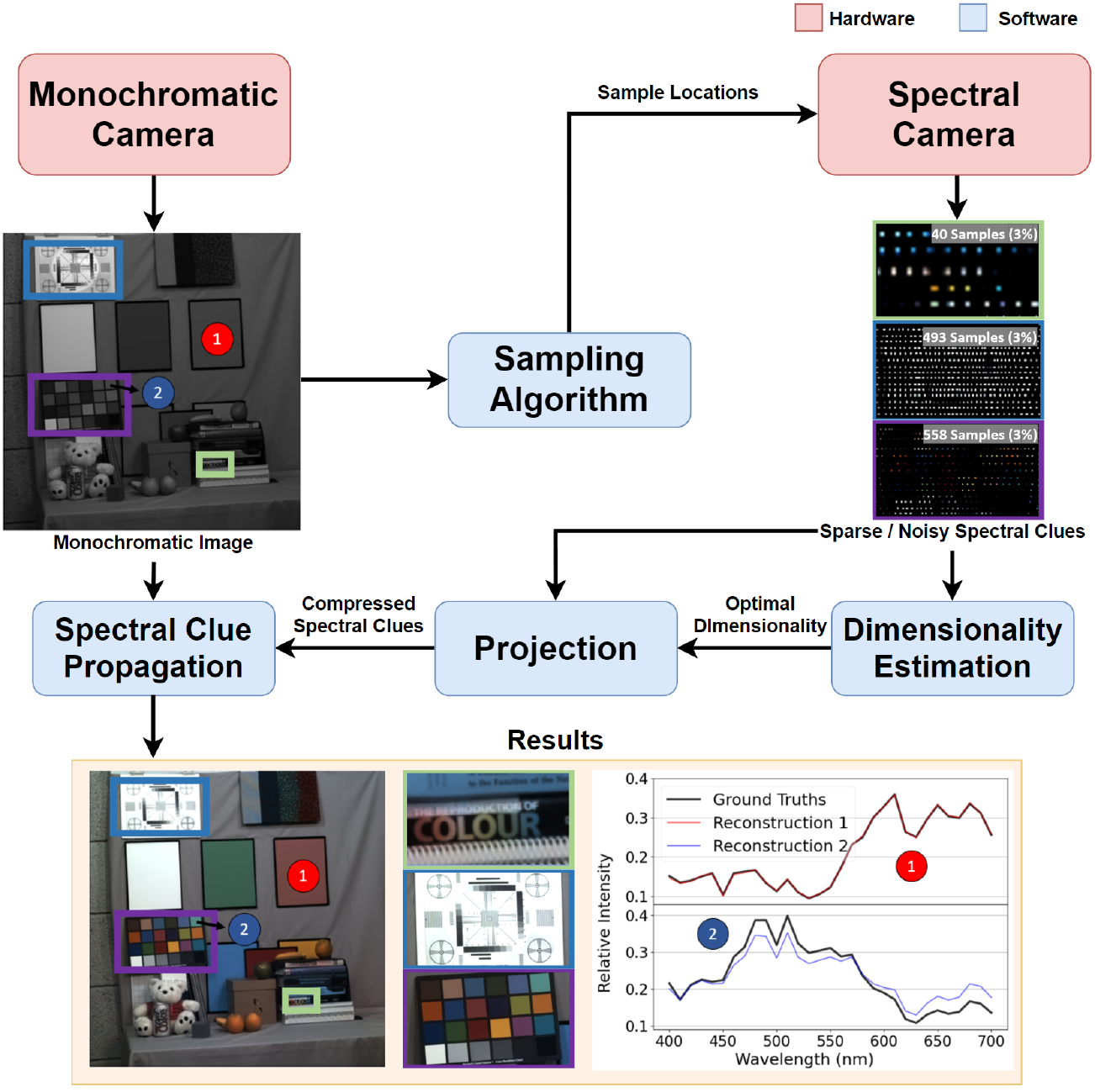}
    \caption{HyperColorization simplified pipeline with an example image from \cite{brainard1998bear}. We use a grayscale image to guide a whisk/push broom scanner or a computational imager to collect spatially sparse spectral samples. These noisy samples are used to estimate the best colorization dimension. Note that we show spatially uniform sampling at a rate of 3\% but can see additional performance gains from image-guided sampling. Finally, the sparse samples are densified using an optimization-based spectral sample propagation algorithm. The code is available at https://github.com/NUBIVlab/HyperColorization and includes demos, documentation of each figure, and spectral results for every pixel in each image shown.}
    \label{fig:intro}
\end{figure}

This paper describes an interpretable, state-of-the-art method to accurately reconstruct a hyperspectral datacube using data captured by hybrid camera systems, where the data is a spatially sparse set of spectral measurements (captured by the spectral camera) and a grayscale image (captured by the supporting camera). By selectively sampling the hyperspectral datacube with the spectral camera, we can achieve significant reductions in capture times when employing traditional push or whisk broom scanners or when using computational imagers, such as the ones described in 
\cite{saragadam2021sassi, cao2011high}. Our method draws inspiration from the work of Levin et al. in \cite{levin2004colorization}, which colorizes grayscale images based on user-drawn color scribbles. We expand upon their assumption that two neighboring pixels should have similar RGB colors if their intensities are similar. We propose that if two neighboring pixels exhibit similar intensities in the grayscale image, their spectral responses, within or beyond the visible spectrum, should also be similar in HSI. Going beyond the conventional wavelength-based binning, where each spectral channel corresponds to a narrow band, our method can be adapted to low-rank representations of spectral data. Performing colorization in a low-rank space is computationally more efficient and has better robustness against noise than colorization in the traditional full-rank representation. We verify that the algorithm's optimal colorization dimensionality depends on the shot noise level and can be estimated using noisy spectral measurements. This allows us to adaptively determine the most suitable dimensionality for accurate reconstruction and effective noise mitigation. Further improvements are gained with guided sampling and edge-aware clue averaging, illustrated in Fig. \ref{fig:intro} and described in detail below. We refer to this framework as HyperColorization (HC).

\begin{figure}[ht!]
    \centering\includegraphics[width=12cm]{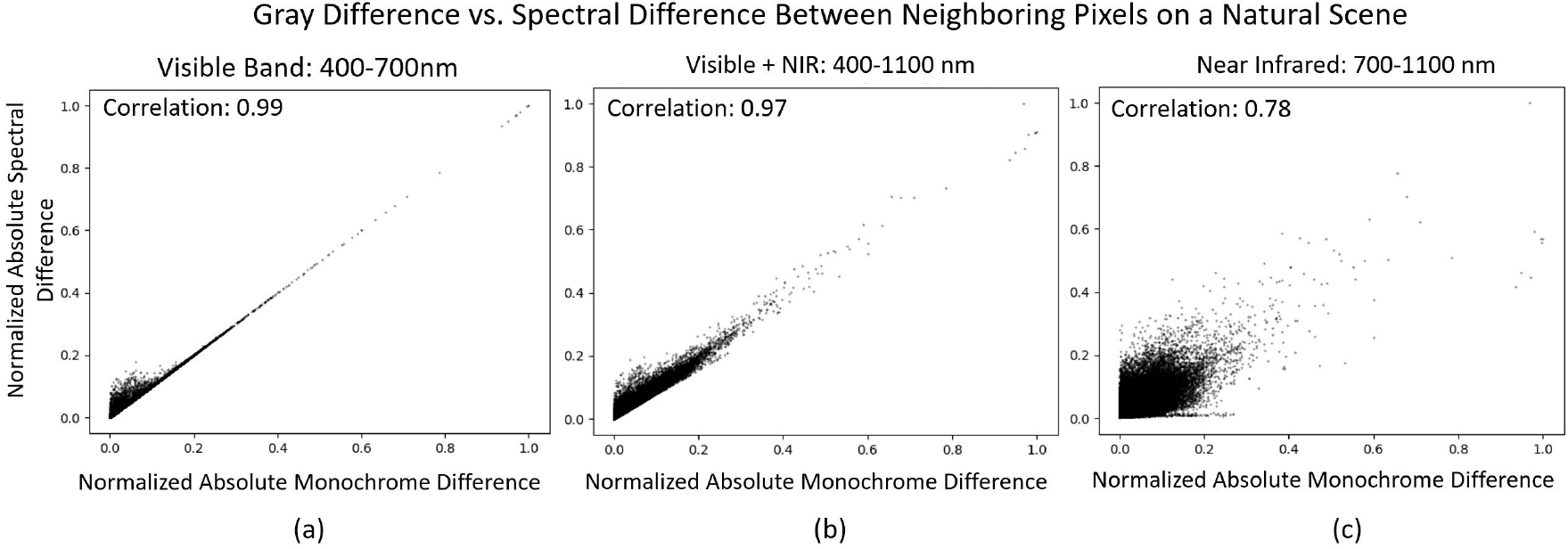}
    \caption{Luminance differences between neighboring pixels carry information about spectral differences. The absolute difference between values on grayscale guide images (simulated by summing across visible spectral channels) is correlated with the L1 distance between their spectra. This correlation is strongest in (a), where the spectral differences are taken along the same wavelength range as the guide image. The relationship weakens in (b) where spectral bands not included in the guide image are considered, and is weaker still in (c) where there is no spectral overlap between the guide image's spectral bands and our target spectral bands. Data shown here is drawn from the dustbin and bulb images of the ICVL dataset \cite{arad2016sparse} which includes 519 bands from 390 to 1043 nm. 
    }
    \label{fig:correlation}
\end{figure}

Our contributions can be summarized as follows:
\setlist{nolistsep}
\begin{itemize}[noitemsep]
    \item {Development of a robust colorization-by-optimization method designed to improve the data fusion process of hybrid camera systems that outperforms the state of the art.}
    \item Proposal of an ideal dimensionality approximation algorithm that estimates the optimal rank of color space for effective colorization from noisy spectral samples.
    \item Introduction of grayscale-image-guided sampling methods that intelligently adjust the sampling frequency for whisk and push broom scanners or suggest sampling locations for computational imaging systems. 
    \item Analysis of the trade-off between the number of measurements and the exposure time for each measurement within a fixed time budget, providing insights into optimizing data collection strategies. 
\end{itemize}
\section{Method}
    \subsection{Hyperspectral colorization using optimization}
        Our color propagation algorithm is based on the assumption that neighboring pixels that have similar intensities have a high probability of sharing similar spectral responses. Our premise is directly analogous to the assumption of Levin et al. in \cite{levin2004colorization}, where they postulate that nearby pixels with similar gray levels should possess similar RGB colors. In Levin's work, color propagation was performed specifically on the U and V channels of the YUV color space. This deliberate choice allowed the propagation of colors while preserving the original luminance information encoded in the Y channel. However, perception-driven hyperspectral and multispectral color spaces are comparatively underdeveloped, limiting generalizability to HSIs. We show in Figure~\ref{fig:correlation} that grayscale guide images generated by summing across spectral channels in the visible range meet this assumption for HSIs, even when the target spectra include or consist entirely of near infrared bands. The correlation of gray level difference between neighboring guide image pixels and the L1 distance between their spectra in Fig.~\ref{fig:correlation}a (correlation: 0.99) confirms that grayscale guide images are informative for visible spectra. The correlation of this visible-based guide image into the near-infrared (NIR) range (400-1100 nm in Fig.~\ref{fig:correlation}b with correlation 0.97 and 700-1100 nm in Fig.~\ref{fig:correlation}c with correlation .78) suggests a promising robustness to the spectral sensitivity used to generate guide images.

        Our algorithm requires a grayscale guide image \(G\in \mathbb{R}^{m \times n}\) aligned with a spatially sparse data cube \(M\in \mathbb{R}^{m\times n\times l}\) that contains measured spectral clues. 
        In the case of hybrid camera systems, data in $M$ is collected by the spectral camera and $G$ is captured by a grayscale camera. To simulate the measurement process, we sample spectral vectors from a HSI to obtain M, and we average the channels that are in the visible range to obtain G. 
        We will use bold \(\textbf{r}\) and \(\textbf{s}\) to indicate 2D pixel coordinates, therefore, \(M(\textbf{r})\) and \(M(\textbf{s})\) represent \(l\) dimensional vectors. To propagate spectral clues, we will individually minimize the criteria described below over all the spectral channels:
        \begin{equation}
            loss(H(:,\lambda))=\mathlarger{\sum_\textbf{r}}\left( \kappa \cdot H(\textbf{r},\lambda) - \frac{\sum_{\textbf{s} \in N(\textbf{r})}w_{\textbf{r,s}}H(\textbf{s}, \lambda)}{\sum_{\textbf{s} \in N(\textbf{r})}w_{\textbf{r,s}}}-M(\textbf{r}, \lambda)\right)
        \label{eq:colorprop}
        \end{equation}
        We minimize this equation over \(H\in \mathbb{R}^{m\times n\times l}\), which serves as the densified version of \(M\). 
        In the equation above, \(N(\textbf{r})\) represents the set of pixels in the n-by-n patch surrounding \(\textbf{r}\) as its neighbors, where n=3 for the results demonstrated in this paper. The value of \(\kappa\) is 2 if \(M(\textbf{r})\) contains a spectral clue; otherwise, it is set to 1. Finally, \(w_{\textbf{r,s}}\) is the weight that encodes the similarity between pixels \(\textbf{r}\) and \(\textbf{s}\) in the grayscale image. We calculate \(w_{\textbf{r,s}}\) using one of the affinity functions described in \cite{levin2004colorization}:
        \begin{equation}
            w_{\textbf{r,s}} \propto e^{-\frac{(G(\textbf{r})-G(\textbf{s}))^2}{2\sigma_{\textbf{r}}^2}}
        \label{eq:affinity}
        \end{equation}
        In Eq. \ref{eq:affinity}, \(\sigma_{\textbf{r}}^2\) is the variance in the patch surrounding \(\textbf{r}\). If \(\textbf{r}\) and \(\textbf{s}\) have similar intensities in the grayscale image \(G\), then \(w_{\textbf{r,s}}\) should assume a higher value to enforce a similar spectral response between the two pixels.
        \begin{figure}[ht!]
            \centering\includegraphics[width=10cm]{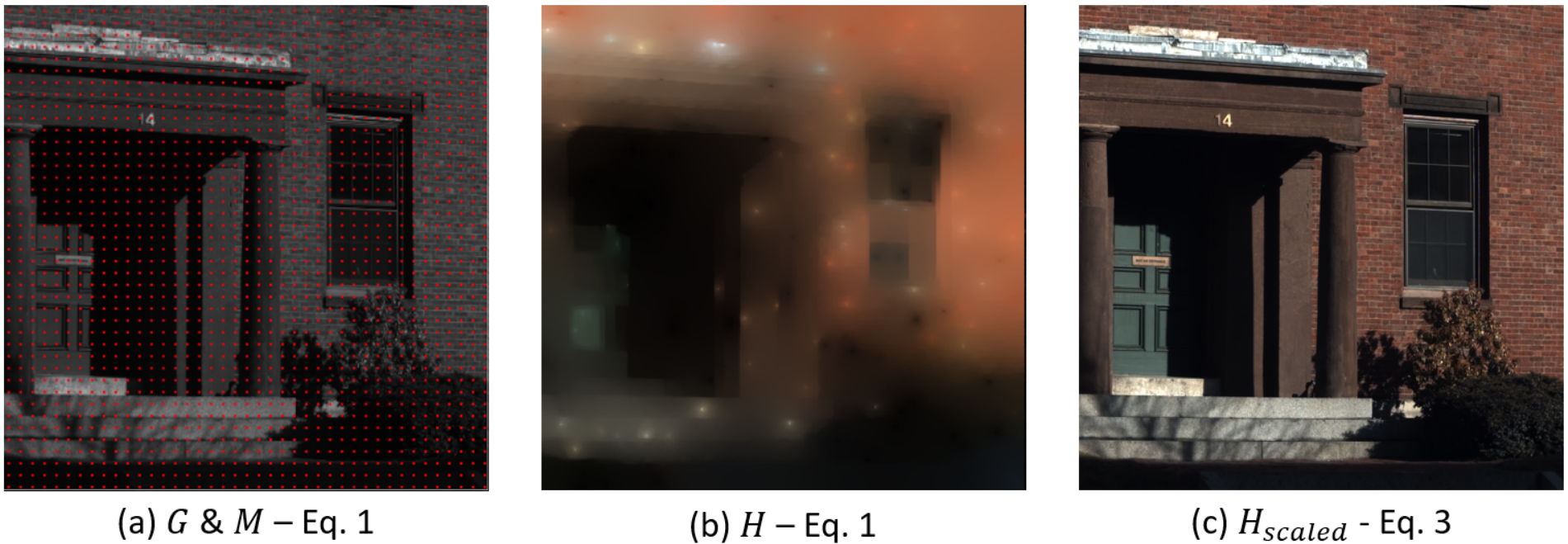}
            \caption{Luminance Adjustment. (a): Grayscale guide image, created by summing across wavelengths, with red dots indicating the uniformly-spaced locations of spectral clues (shown over 2\(\times\)2 pixels for visibility). (b): Color propagation does not preserve luminance information on hyperspectral images. (c): Renormalization of image (b) based on image (a) using Eq. \ref{eq:lumbalance} removes these artifacts. Image from \cite{chakrabarti2011statistics} contains 31 channels from 400 to 700 nm and has been projected to RGB here for visualization.}
            \label{fig:lumbalance}
        \end{figure}
        
        Propagating spectral clues in this manner causes luminance information to be lost in the output, as HSIs don't have a dedicated Y channel to preserve this information as in the YUV color space. As a result, errors in overall brightness can arise across local patches in the image (Fig. \ref{fig:lumbalance}b). However, the relative intensities in spectral response vectors remain accurate, so the problem can be fixed by normalizing the spectral vectors according to the grayscale image, using Eq. \ref{eq:lumbalance}:
        \begin{equation}
            H_{scaled}(\textbf{r},\lambda )=\alpha \cdot \frac{G(\textbf{r})}{\sum_{\textbf{r}}|H(\textbf{r},\lambda )|\cdot \frac{S_G(\lambda )}{S_H(\lambda )}}H(\textbf{r},\lambda )
        \label{eq:lumbalance}
        \end{equation}
        Where \(\alpha\) is a camera-pair-dependent parameter and \(S_G(\lambda )\) and \(S_H(\lambda )\) are normalized spectral response functions of the grayscale and spectral cameras. This equation assumes a linear relationship between the L1 norm of the spectral vector and the intensity of the grayscale image. 
        Fig. \ref{fig:lumbalance}a shows the simulated guide image that is used to renormalize the original reconstruction in panel b (31 channels projected to RGB for visualization), producing the corrected reconstruction result shown in panel c.

        Traditional methods like bilateral or edge-aware filtering are popular in handling noisy images. However, applying a bilateral filter after color propagation is ineffective because propagation spreads each spectral sample across a large area, creating spatially consistent noise. Instead, a better approach is to average spatially close spectral samples before color propagation. We perform this averaging with edge-aware filtering to prevent color bleed:
        \begin{equation}
            M_{filtered}(\textbf{r},\lambda )=\zeta \cdot M(\textbf{r},\lambda ) + \frac{1-\zeta }{|N(\textbf{r})|}\cdot
            \sum_{\textbf{s} \in N(\textbf{r})}M(\textbf{s},\lambda ).
        \label{eq:smartfilter}
        \end{equation}
        Here, \(\zeta \in [0,1]\) is the blurred result of Canny edge detection on the grayscale image with a normalized Gaussian filter; it acts as a weight parameter that captures the spatial proximity of \(\textbf{r}\) to an edge, and \(N(\textbf{r})\) is the neighborhood of spectral measurements in the vicinity of \(\textbf{r}\), which includes the nearest \(|N(\textbf{r})|\) pixels. We show results using a \(21 \times 21\) pixel neighborhood and the Gaussian kernel is \(31 \times 31\) with variance 11.
        
    \subsection{Hyperspectral colorization in a learned subspace}
\begin{figure}[ht!]
            \centering\includegraphics[width=12cm]{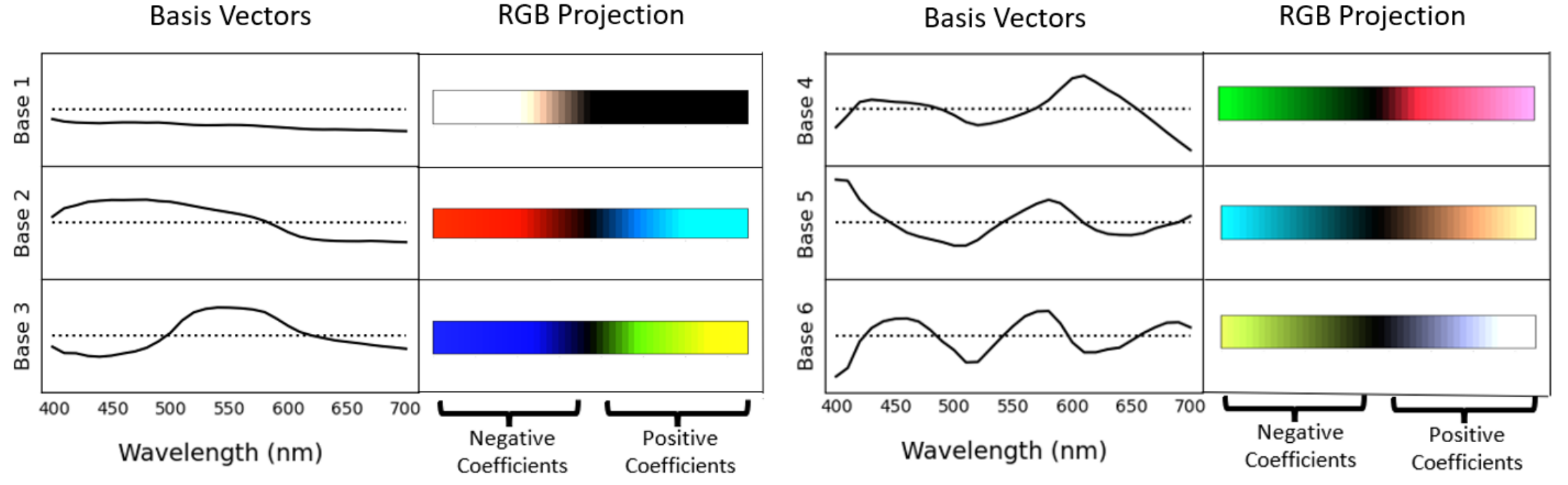}
            \caption{Singular vectors corresponding to the largest 6 singular values learned from the CAVE dataset \cite{yasuma2010generalized} and their RGB  projections for coefficients from -15 to +15. Note that RGB channels have been allowed to saturate.}
            \label{fig:basevecs}
        \end{figure}

        It is well known that the collection of distinct spectral responses in a natural image occupy a low-rank space, and there are many approaches to represent HSIs using low dimensional models \cite{rasti2013hyperspectral, li2011compressive,chakrabarti2011statistics,lee2000spectral}. In this paper, we will follow the results of Chakrabarti et al. in \cite{chakrabarti2011statistics} and Lee et al. in \cite{lee2000spectral}, to treat a hyperspectral image as a 2D matrix, \(H^{r \times l}\), where \(r\) is the number of pixels and \(l\) is the dimension of the spectral vector of each pixel. By applying singular value decomposition (SVD), we can extract \(l\) singular vectors that span \(\mathbb{R}^l\). Fig. \ref{fig:basevecs} displays the singular vectors corresponding to the most significant singular values learned from the CAVE dataset \cite{yasuma2010generalized}. Treating these singular vectors as an orthonormal basis, we can project our hyperspectral image onto the top \(p<l\) vectors, effectively reducing the number of channels. This compressed representation of the hyperspectral data can be viewed as a learned color space, and we will show that the HyperColorization algorithm can work without loss of fidelity in the low dimensional space, given that a proper value for \(p\) is chosen.

        In the presence of noise, the choice of dimensionality \(p\) of HyperColorization becomes critical to performance. By propagating spectral clues in a lower dimension, we not only reduce computational costs but also remove noise components that are orthogonal to the subspace spanned by the selected vectors, thereby increasing the signal-to-noise ratio (SNR). However, if \(p\) is chosen too small, singular vectors will underfit the data and cause a drop in HyperColorization accuracy. On the other hand, if \(p\) is chosen to be too large, the projection may “overfit” to the measurement, preserving shot noise. The optimal reconstruction dimensionality depends on the intrinsic dimensionality of the scene and the level of noise present. Accurately estimating this dimensionality before colorization is crucial.

\begin{figure}[ht!]
            \centering\includegraphics[width=8cm]{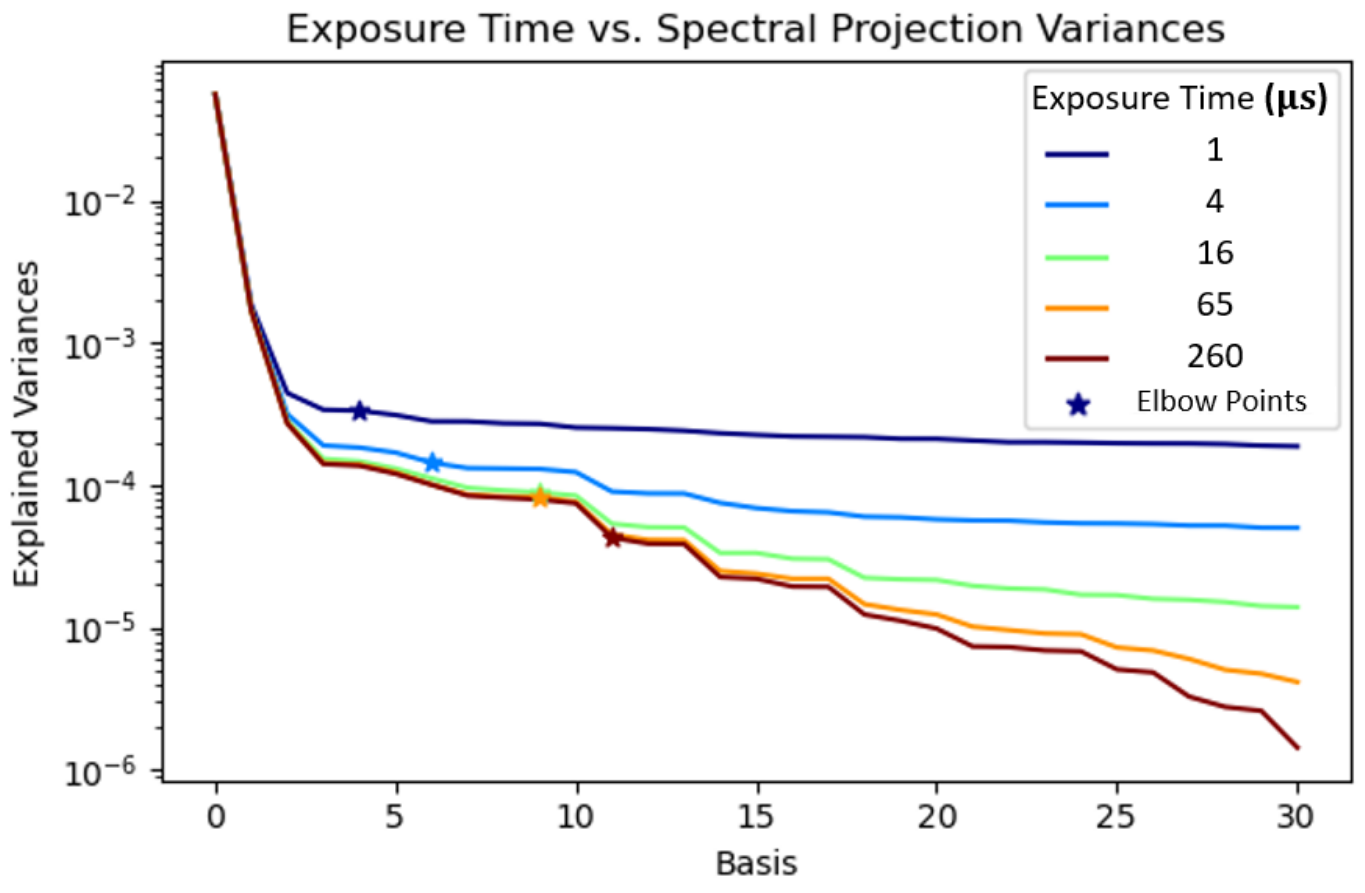}
            \caption{The relationship between noise level and the projection of hyperspectral clues onto our learned basis. The elbow location and vertical offset of these curves vary with noise and can be used to predict the highest-performing reconstruction dimensionality.}
            \label{fig:obs}
        \end{figure}

        To estimate the optimal reconstruction dimensionality, we have devised a method that leverages noisy spectral samples and our basis vectors. As depicted in Fig. \ref{fig:obs}, we make the following observations: (1) As the noise levels increase, the minimum variance of projections onto our learned color channels also tends to increase. (2) The location of the ‘elbow’ points, as defined in \cite{satopaa2011finding}, gradually shifts towards lower dimensions with greater noise. Both phenomena can be attributed to the white characteristic of Poisson noise, which evenly distributes the noise power across dimensions. These two features (minimum and elbow point of explained variance curve) can be used to predict the reconstruction dimension of highest performance.

\begin{figure}[ht!]
            \centering\includegraphics[width=12cm]{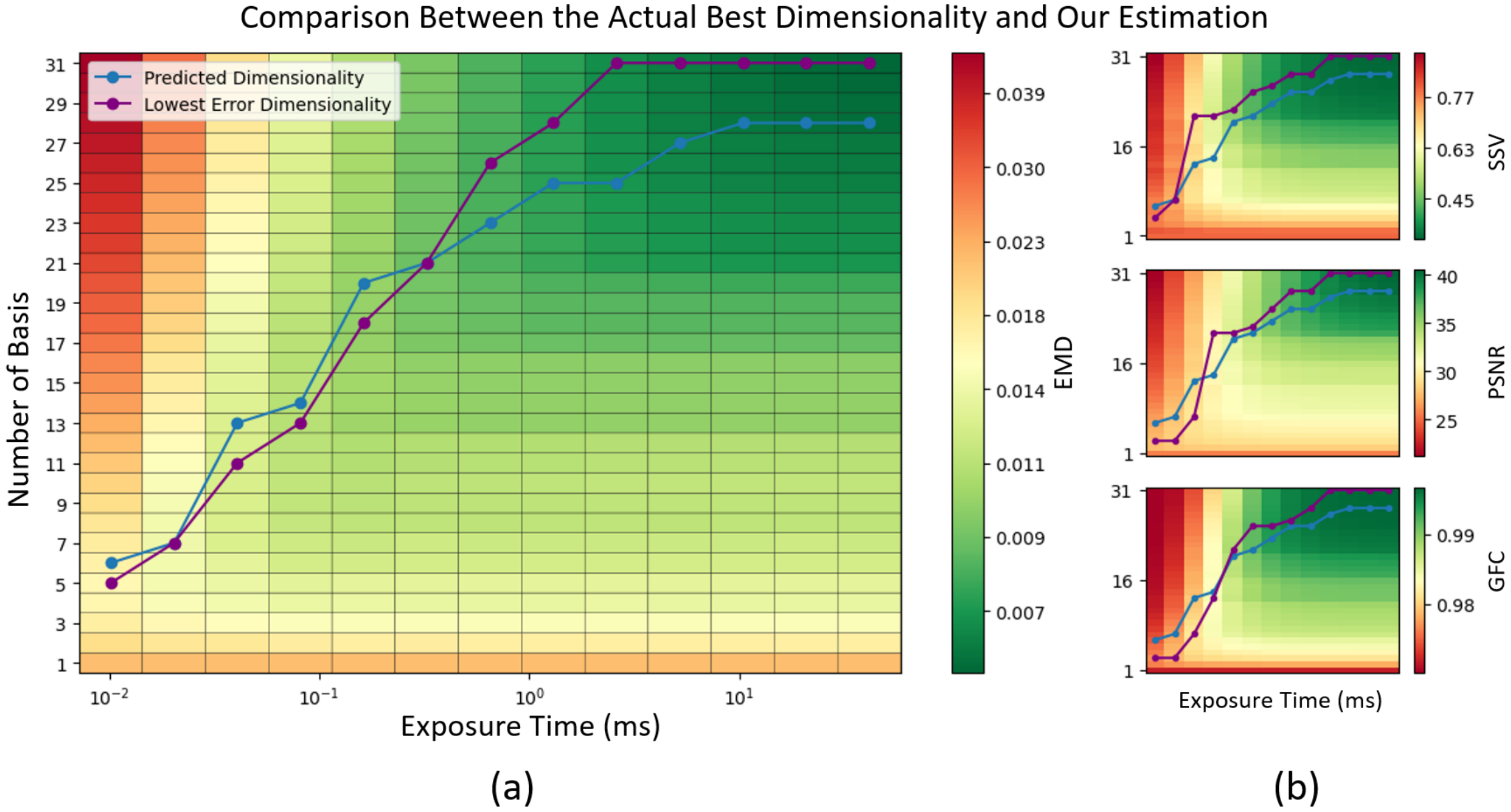}
            \caption{Reconstruction error across metrics, exposures times, and reconstruction dimensions for the sample Bear \& Fruit image. (a) We propose EMD as a useful metric for hyperspectral quality due to the well-behaved purple curve. Generalizing to the full CAVE dataset, we can approximately predict (blue) this curve using the features described in Fig. \ref{fig:obs}. (b) Other error metrics show sudden jumps and plateaus in their best-performing dimension, and suggest very low dimensions at short exposure times.}
            \label{fig:heatmap}
        \end{figure}

        In Fig. \ref{fig:heatmap}, we performed a grid search to determine the best colorization dimensionality at each exposure time over the image from~\cite{brainard1998bear} shown in Fig \ref{fig:intro}. To quantify the reconstruction quality, we used earth movers distance (EMD) because compared to other quality metrics, also shown in Fig. \ref{fig:heatmap}, it offers two advantages. First, it scales more linearly on log-scale of exposure time, avoiding abrupt transitions and plateaus  (compare purple lines across subfigures). Secondly, it retains spectral diversity even under extreme noise (compare 5 vs. 2 basis vectors at lowest exposure time). 
        Generalizing from this example (see Fig. S1-3 for more), we trained a second-order polynomial regression model to predict the best-EMD reconstruction dimension using the CAVE dataset \cite{yasuma2010generalized}. The predicted dimensionality is shown in blue and repeated across subfigures.
        

\begin{figure}[ht!]
            \centering\includegraphics[width=12cm]{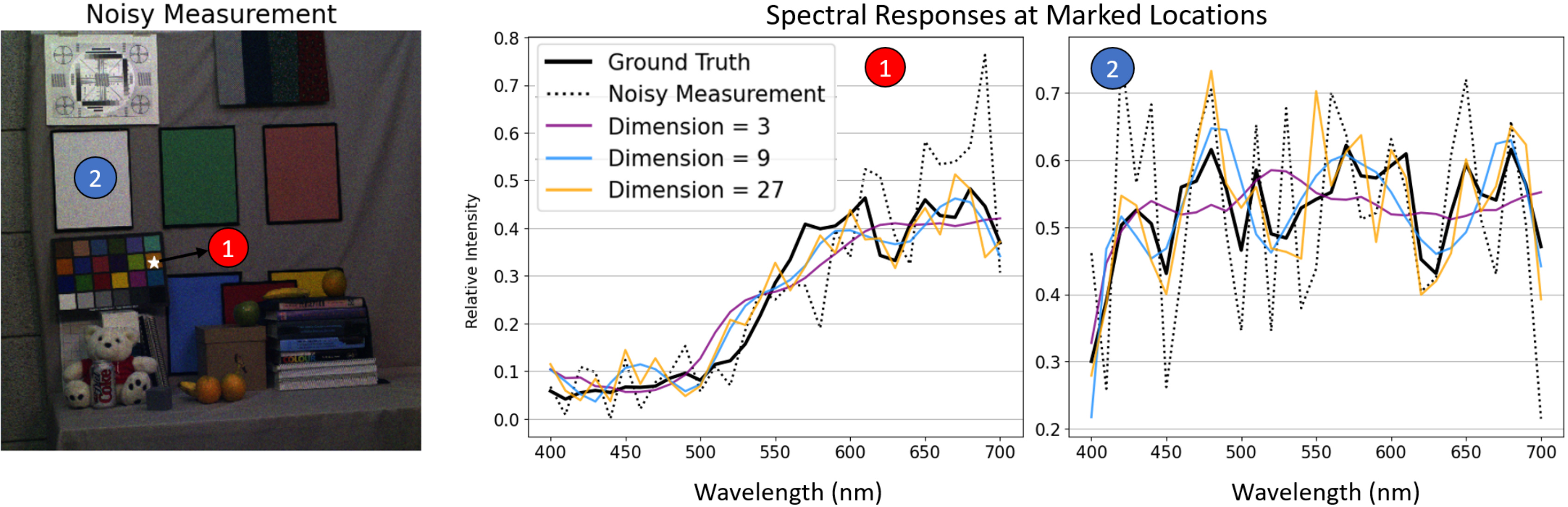}
            \caption{The impact of reconstruction dimension with 1\% sampling ratio. In this example, 3-dimensional reconstruction underfits to the data, while 27-dimensional reconstruction overfits to noise. The optimal reconstruction dimensionality is 9. Image\cite{brainard1998bear} contains 31 visible bands, projected onto RGB for visualization.}
            \label{fig:spectralcurves}
        \end{figure}

        Fig. \ref{fig:spectralcurves} illustrates the effects of reconstruction dimensionality on samples with a simulated exposure time of 40\(\mu\)s per pixel. A high-dimensional reconstruction (27D, shown in yellow) overfits to the noise spikes seen in the measurement (dotted black) while a low dimensional reconstruction (3D, shown in magenta) underfits the data and excessively smooths the spectra. The correct dimension (9D, shown in blue) more closely matches the ground truth spectra (solid black). 
        
    \subsection{Grayscale guided sampling for push and whisk broom cameras}
        Images of natural scenes often feature large patches with low-frequency brightness changes and consistent color spectra, alongside localized image patches with high-frequency spatial details and more diverse color spectra. When a grayscale image of the scene is available before spectral sampling, spatial sampling patterns can be adapted to prioritize high-variance patches.

        During color propagation, spectral samples are competing to fill the surrounding region. When the sampling algorithm is uniform, the distance between the samples is equal, ensuring a fair competition between the samples. If our sampling algorithm does not preserve this uniformity as much as possible, some samples may overflow their true extent by not getting challenged by other samples. Consequently, segmentation based sampling algorithms, such as the one discussed in \cite{saragadam2021sassi}, are not well-suited for our color propagation algorithm.
        
\begin{figure}[ht!]
            \centering\includegraphics[width=12cm]{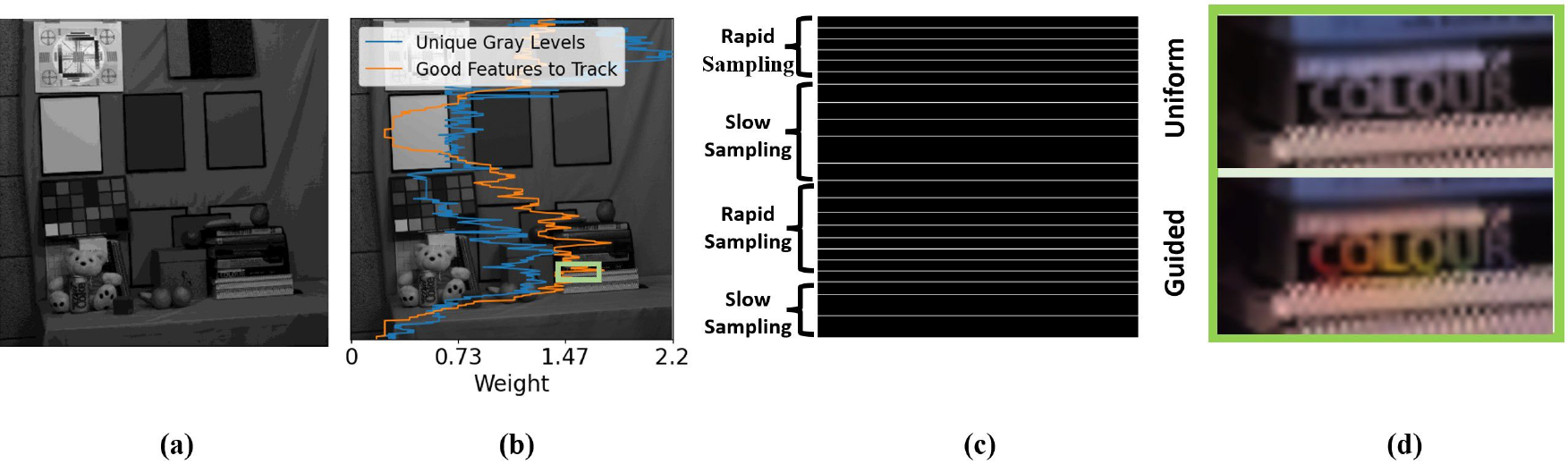}
            \caption{Guided sampling improves performance. (a) A posterized guide  image from \cite{brainard1998bear} is shown with line-by-line scores for the features that guide sampling (b). (c) Our adaptive algorithm adjusts the sampling ratio so that the regions expected to have higher color diversity have a higher sample density. (d) An example inset where uniform sampling fails to capture enough spectral samples to colorize accurately. 31 channel data is projected onto RGB for visualization.}
            \label{fig:spresult}
        \end{figure}

        Our adaptive sampling algorithm considers the diversity of gray levels and local spatial features to determine if a local patch in the image might be rich in spectral characteristics. For uniform sampling with a push broom scanner, we regularly sample every \(n=1/sampling\, rate\) row, giving equal weight to each row. In contrast, our adaptive sampling method for push broom scanners assigns a weight to each row based on two features. The first feature is the number of unique gray levels close by in the reduced-bit-depth representation of the grayscale image. The second feature is the number of ‘good features to track’ \cite{shi1994good} in the vicinity in the grayscale image. We scale and normalize these features so that the minimum value is approximately 0.1 and the mean is 1. Finally, we combine the  features using weighted averaging. As we traverse the image, we accumulate the weight; and take a measurement when the value surpasses the threshold of \(n\). Weights are adjusted according to:
        \begin{equation}
        \begin{aligned}
            f_{\textit{sampling}} \propto ~ \alpha \cdot \Gamma(\textit{\# Good Features to Track}) + (1-\alpha) \cdot \Gamma(\textit{\# Unique Gray Levels}), \\
            \text{where~~}\Gamma(x):= \eta\left(0.1+0.9 \frac{x-\min(x)}{\max(x)-\min(x)}\right), ~~~~~~~~~~~~~~~~~~~~~~~~~~~~~~~~~~~~\\
            \eta(x) := \frac{x}{\text{mean}(x)},~~~~~~~~~~~~~~~~~~~~~~~~~~~~~~~~~~~~~~~~~~~~~~~~~~~~~~~~~~~~~~~~~~~~~~~~
            \label{eq:guided}
        \end{aligned}   
        \end{equation}
        with $\alpha=.7$ for the results reported in this paper based on a grid search. Figure~\ref{fig:spresult}a shows a posterized guide image with weights shown in panel b and guided sample locations in c. The insets in Fig.~\ref{fig:spresult}d show an example location where guided sampling leads to a more accurate reconstruction compared to uniform sampling. Additional sampling examples for push and whisk broom sensors can be seen in Fig. S4-6.
    
        
        The adaptive sampling method for whisk broom scanners extends the algorithm for push broom scanners. Once the rows for sampling are chosen, we assign weights to each pixel along those selected rows based on the two features previously described. We accumulate the weights on each pixel and take a measurement when it surpasses a threshold determined by the desired sampling rate. 
        

    \begin{table}
        \centering
        \caption{Datasets used for simulations.}
        \begin{tabular}{lccc}
        \hline
            \textbf{Dataset} & \textbf{Spatial resolution} & \textbf{Spectral channels} & \textbf{Spectral range (nm)}\\
        \hline
            Harvard \cite{chakrabarti2011statistics} & 1392 \(\times\) 1040 & 31 & 420-720\\
            KAIST \cite{choi2017high} & 2704 \(\times\) 3376 & 31 & 420-720\\
            Bear \& Fruit \cite{brainard1998bear}& 2020 \(\times\) 2020 & 31 & 400-700\\
            CAVE \cite{yasuma2010generalized}& 512 \(\times\) 512 & 31 & 400-700\\
            ICVL \cite{arad2016sparse} & 1300 \(\times\) 1392 & 519 & 390-1043\\
        \hline
        \end{tabular}
        \label{tab:datasets}
    \end{table}
    
\section{Results}
    We tested our algorithm in simulation using the five datasets in Table \ref{tab:datasets}, and assume the Poisson-Gaussian noise model when simulating the grayscale image and the HSI. Specifically, the value at each image pixel of the simulated grayscale image $G(\mathbf{r})$ is
    \begin{align}
        G(\mathbf{r}) = \frac{1}{\rho\cdot t} \left(\text{Poisson}\left(\rho\cdot t \cdot \sum_\lambda H^*(\mathbf{r},\lambda)\right) + \text{Gaussian}(\mu, \sigma^2)\right),
    \end{align}
    and the pixel value of the simulated HSI image $H(\mathbf{r}, \lambda)$ is
    \begin{align}
        H(\mathbf{r},\lambda) = \frac{1}{\rho\cdot t} \left(\text{Poisson}\left(\rho\cdot t\cdot H^*(\mathbf{r},\lambda)\right) + \text{Gaussian}(\mu, \sigma^2)\right),
    \end{align}
    where $H^*(\mathbf{r},\lambda))$ is the HSI measurement from the dataset that we use as ground truth. The coefficient $\rho$ is the brightness adjustment factor, which we set as $9.6\times 10^7$ photons per full grayscale range. This approximately corresponds to the brightness level of an office setting~\cite{EN12464-1} using a FLIR grayscale camera~\cite{camera-datasheet}. Exposure time $t$ was set by the experiment. Finally, we set the parameters of the Gaussian noise to be $\mu = 0$ and $\sigma=0.1$. 

    
    
    The metrics we report in our experiments include peak signal-to-noise ratio (PSNR), goodness factor coefficient (GFC), spectral similarity value (SSV), and SSIM \cite{shrestha2014quality}. 
    Additionally, we propose using earth-mover’s distance (EMD) \cite{ramdas2017wasserstein} or “Wasserstein distance” to compare ground truth spectra and reconstruction spectra at each pixel and average it through the image. This generalization of KL divergence provides a useful assessment of hyperspectral accuracy, particularly when spatial reconstruction errors are negligible. 
    
    \subsection{Sampling patterns}
        We conducted experiments using random, uniform, and guided sampling patterns for both push and whisk broom scanners. Table \ref{tab:samplingpatterns} compares the performance of uniform and guided sampling patterns at a fixed sampling ratio, see Table S1 to compare results of all sampling patterns at various sampling ratios. 
        \begin{table}
            \centering
            \caption{Effect of sampling(4\%) pattern on performance, Harvard dataset \cite{chakrabarti2011statistics}}
            \resizebox{\linewidth}{!}{%
            \begin{tblr}{
              width = \linewidth,
              colspec = {Q[250]Q[90]Q[90]Q[90]Q[122]Q[90]},
              column{5} = {c},
              hline{1-2,6} = {-}{},
            }
            \textbf{Sampling type} & \textbf{PSNR$\uparrow$} & \textbf{SSV\textbf{$\downarrow$}} & \textbf{GFC\textbf{$\uparrow$}} & \textbf{EMD\textbf{$\downarrow \times 10^3$}} & \textbf{SSIM\textbf{$\uparrow$}} \\
            Uniform push broom     & 35.155                  & 0.433                             & 0.989                             & 9.340                                         & 0.9681                           \\
            Guided push broom      & 35.500                  & 0.409                             & 0.989                             & 9.261                                         & 0.9662                           \\
            Uniform whisk broom    & 37.895                  & 0.358                             & 0.992                             & 6.789                                         & 0.9788                           \\
            Guided whisk broom     & 38.457                  & 0.352                             & 0.992                             & 6.607                                         & 0.9803                           
            \end{tblr}
            }
            \label{tab:samplingpatterns}
        \end{table}
    \FloatBarrier
    \subsection{Imaging with a fixed time budget}

\begin{figure}[ht!]
        \centering\includegraphics[width=12cm]{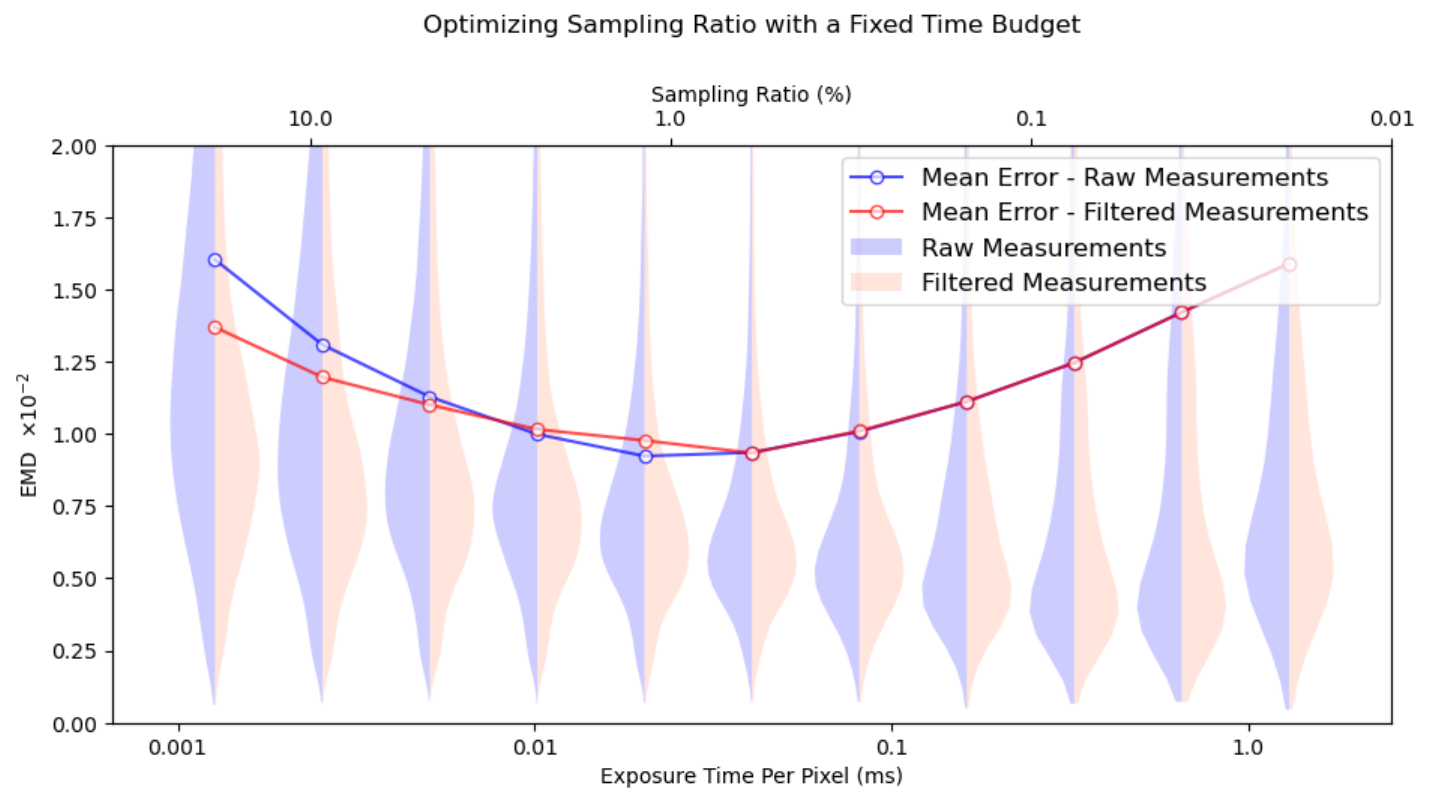}
            \caption{HyperColorization has a per-image optimal sampling ratio and favors oversampling. We show here the histograms of errors on 400 \(\times\) 400 pixel scale Bear \& Fruit image \cite{brainard1998bear} with a total time budget of 5 seconds under guided whisk broom sampling. Notice that as our sampling ratio (top x-axis) increases, our exposure time per pixel (bottom x-axis) decreases. Violin plots represent the distribution of error across pixels. While the optimal sampling ratio varies from image to image, spatial undersampling is a more serious problem than the increased photon noise due to spatial oversampling.}
            \label{fig:timebudget}
        \end{figure}

        When utilizing scanner-type imaging modalities within a limited timeframe, a balance must be struck between the number of samples acquired and the exposure time per sample. Increasing the sampling ratio improves the ability to capture finer details but leads to higher Poisson noise per sample. This is what allows us to break the traditional trade-offs seen in these scanning modalities, by lowering noise on a subset of spectral samples and regaining spatial resolution through the guide image. Table 3 shows the advantage in reconstructing HSIs fom 10\% of pixels rather than using the same total exposure times across all pixels. Conversely, reducing the number of samples risks missing colored regions entirely. Our framework does not inherently determine the optimal sampling ratio given a specific time budget and scene structure. However, based on insights from our data, we provide a practical guideline, as depicted in Figure \ref{fig:timebudget}.

        Examining Fig. \ref{fig:timebudget} reveals that the best strategy is to lean towards capturing more samples at the expense of having more shot noise per sample because our framework offers several tools to deal with noise, including techniques like dimensionality reduction and edge-aware filtering. However, it's crucial to acknowledge a fundamental limitation: colors of regions that were never sampled cannot be recovered. Undersampling an image is only justifiable when our primary interest lies in preserving lower-frequency regions. The presence of the bulge at the lower end of the violin plots indicates that, even in cases where the image is undersampled, most of its content remains accurately colorized. These trends are confirmed on images from Harvard, KAIST and CAVE datasets shown in Fig. S7-9.
    \FloatBarrier
    \subsection{Comparison to the state of the art}
        In our study, we conducted a comparison involving HyperColorization and other hybrid camera systems, along with a snapshot technique. Among the hybrid techniques, SASSI \cite{saragadam2021sassi} and Cao et al. \cite{cao2011high} utilize RGB cameras for guidance and grayscale sensors with dispersing prisms for spectral sample collection. To isolate the measurements on the image sensor, SASSI employs spatial light modulators (SLM) while Cao et al. use a static mask. They later densify the collected spectral samples using their respective algorithms with the guidance of the captured RGB image. 
        
\begin{figure}[ht!]
            \centering\includegraphics[width=12cm]{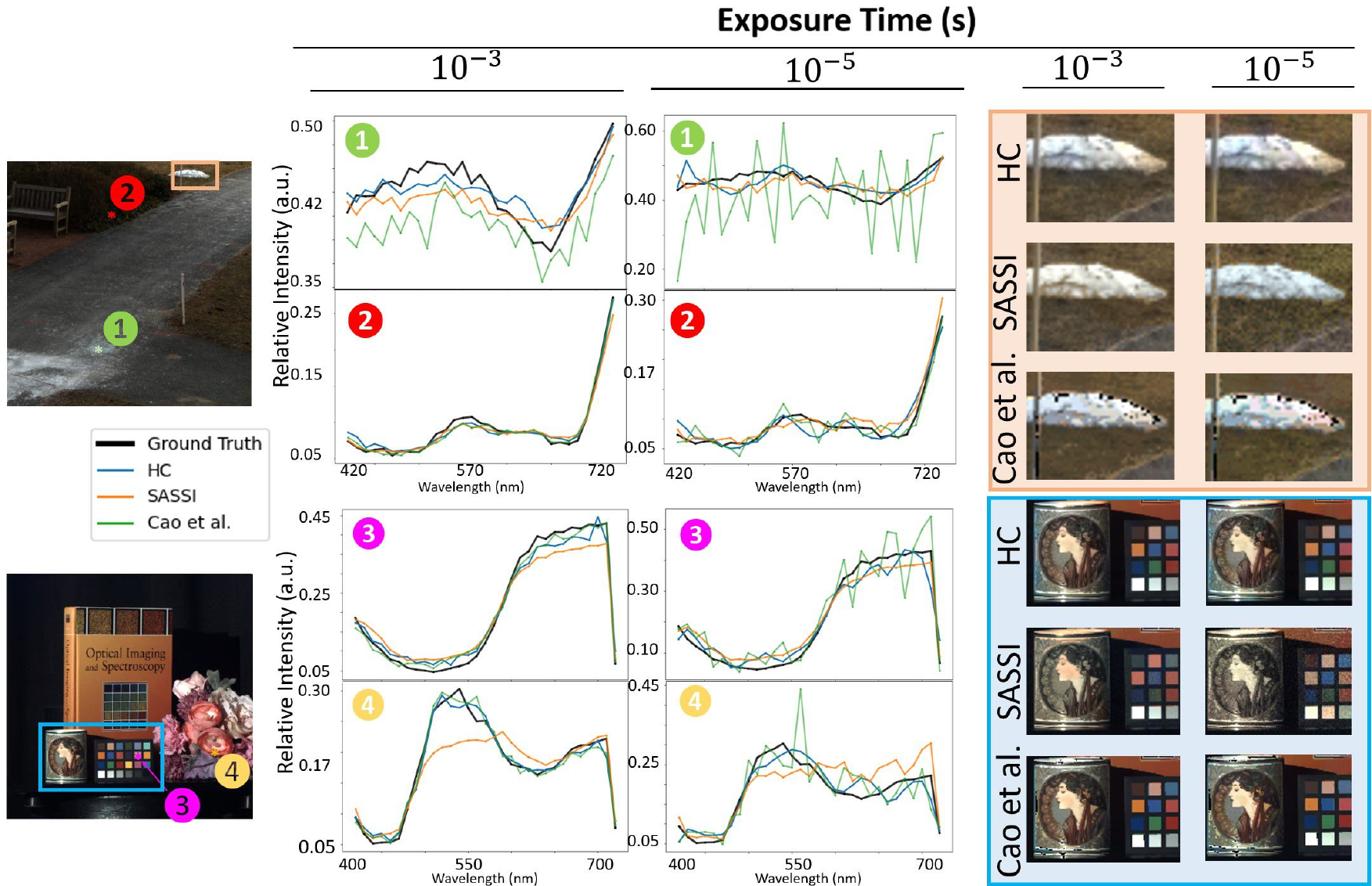}
            \caption{Reconstruction results on example images in the visible range from Harvard~\cite{chakrabarti2011statistics}(top) and KAIST~\cite{choi2017high} (bottom) datasets for two different noise levels simulating different exposure times. We compare to two hybrid architecture methods, SASSI and Cao et al., with parameters set to maximize PSNR (SASSI rank 1, Cao et al. filter sizes 31 with spatial variance of 21 and spectral variance of 0.03, set by grid search) but still result in noisy and desaturated spectra and spatial artifacts such noise and pixelation (orange insets) or color blotches and edge artifacts (blue insets).
            }
            \label{fig:comphybridvisible}
        \end{figure}

\begin{figure}[ht!]
            \centering\includegraphics[width=12cm]{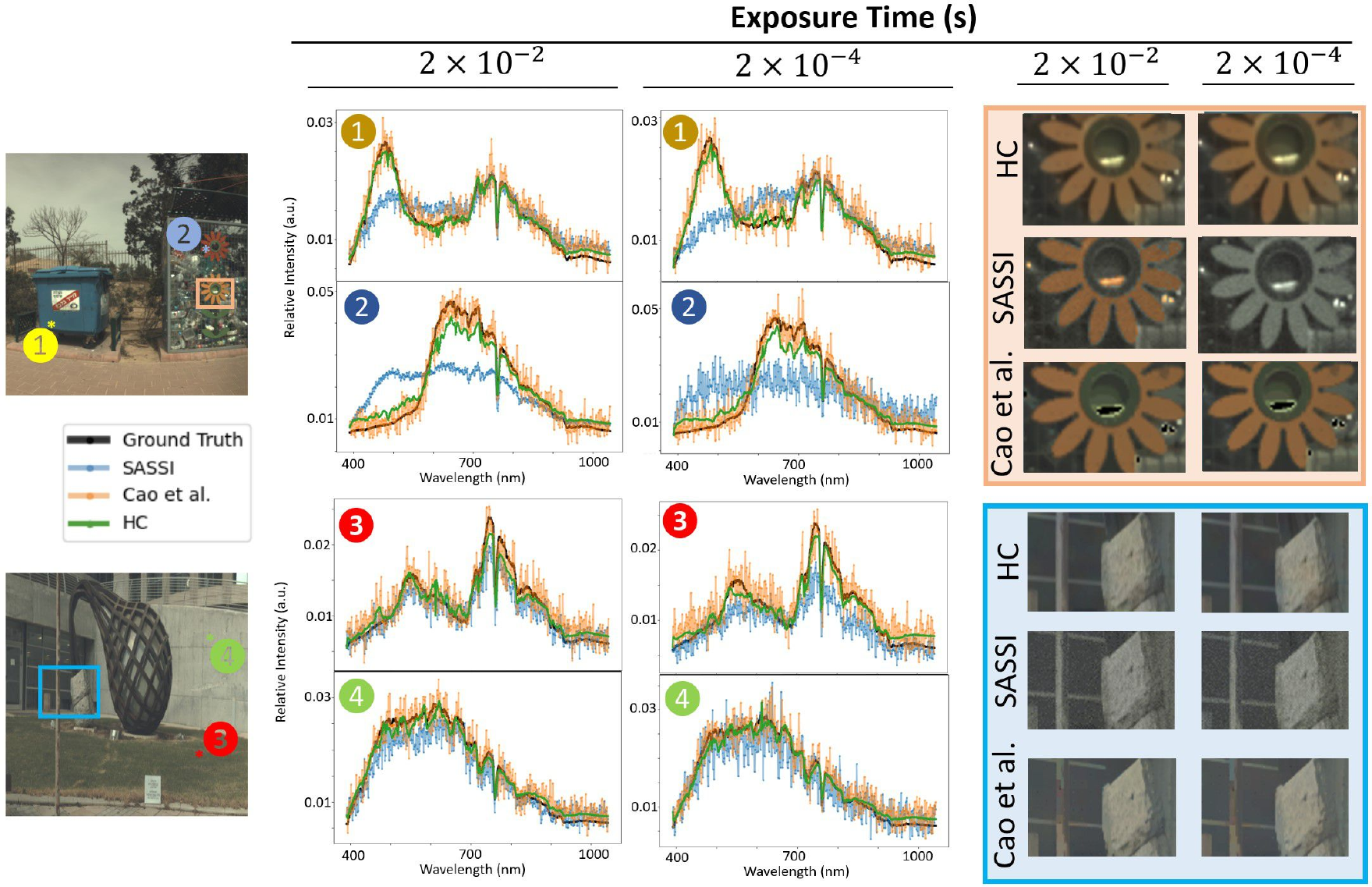}
            \caption{Reconstruction results of hybrid architectures over two example images from ICVL dataset with 3\% guided sampling. These examples have 519 channels in visible and NIR. HyperColorization results are obtained by projecting the noisy images to the top 31 singular vectors trained from the same images with guide images generated from the full spectral range including NIR, conditions likely to provide an upper bound on HC performance at this task. Hyperparameters of SASSI are left at their default values, and hyperparameters of Cao et al. are initialized after as the same values reported in Fig. \ref{fig:comphybridvisible}. Noise suppression causes SASSI to lose saturation, while Cao et al.'s results exhibit spatial artifacts.}
            \label{fig:comphybridIR}
        \end{figure}

\begin{figure}[ht!]
        \centering\includegraphics[width=12cm]{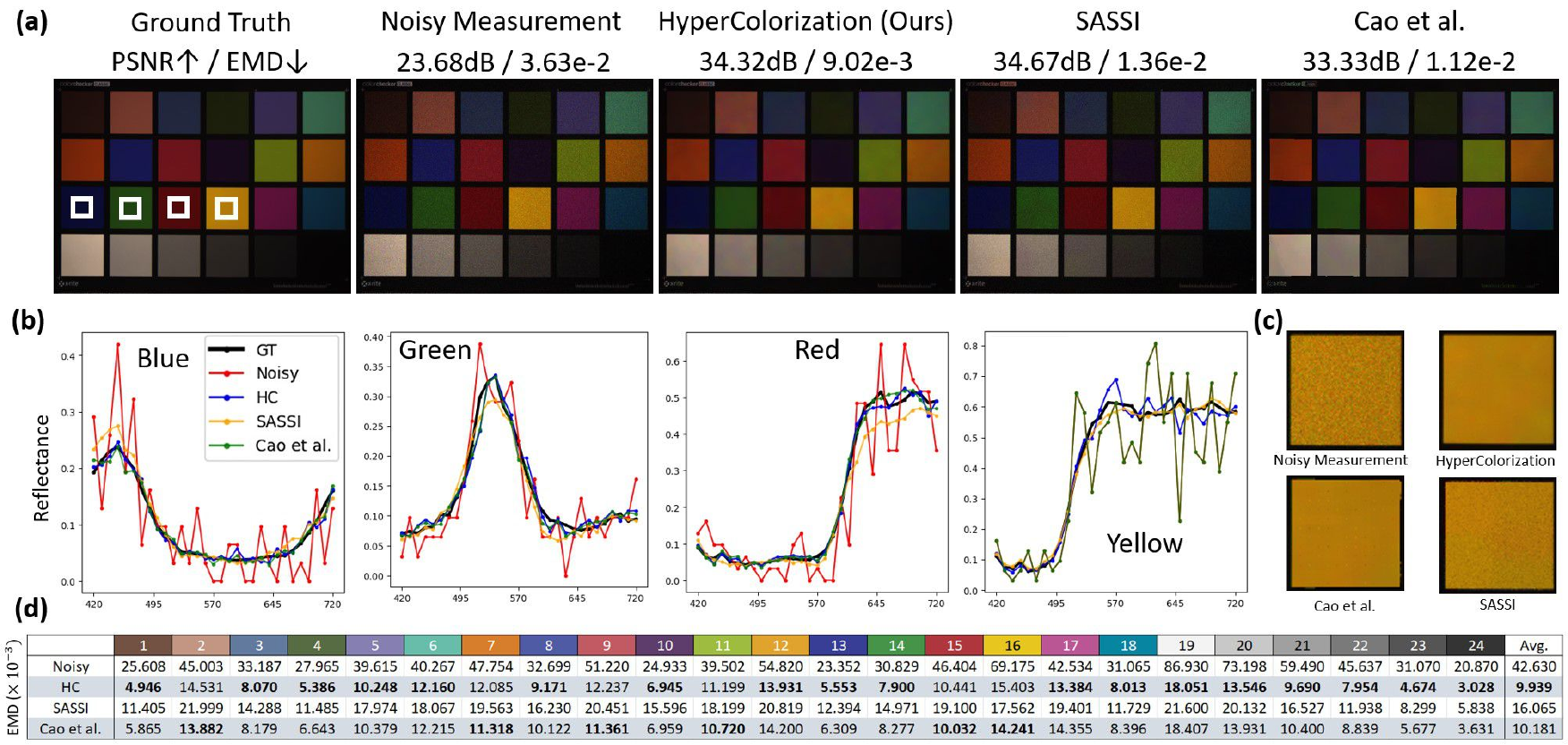}
            \caption{Comparison of hybrid camera systems on a 31-channel, 420-720 nm image of a color checker from the KAIST dataset \cite{choi2017high}. (a) Projections onto RGB show spatial noise reduction in HyperColorization, with PSNR and EMD reported for hyperspectral data over all pixels. (b) Spectra from sample patches, showing  noisy measurements in the yellow patch for Cao et al.'s because their algorithm does not modify the initial color samples. (c) These noisy locations are visible across methods in the yellow patch inset. See Figure S10 for spectral results in all patches, which are summarized numerically in (d).}
            \label{fig:colorchecker}
        \end{figure}

        HyperColorization is designed to densify spectral samples from arbitrary imaging modalities. Besides whisk and push broom scanners, our algorithm can be seamlessly integrated into hybrid spectral imagers to enhance their color propagation accuracy. The results presented in Table \ref{tab:comparisontable} demonstrate that HyperColorization outperforms the spatial densification algorithms employed by SASSI and Cao et al. while maintaining a lower computational cost compared to Cao et al. Additionally, see Fig. \ref{fig:comphybridvisible} to compare reconstructions of hybrid architectures over two images from Harvard and KAIST datasets over the visible band; and see Fig. \ref{fig:comphybridIR} to compare reconstructions  over two images from the ICVL dataset over visible and NIR bands. {Finally, see Figs.~\ref{fig:colorchecker} and S10 to compare results on a color checker.}
        \begin{table}
            \centering
            \caption{Comparison of HyperColorization to classic and hybrid camera systems using 21 images selected from Harvard, KAIST and CAVE datasets.}
            \resizebox{0.88\linewidth}{!}{%
            {\tiny
            \begin{tblr}{
              colspec = {Q[35]Q[1]Q[12]Q[12]Q[12]Q[1]Q[12]Q[12]Q[12]Q[1]Q[13]Q[13]},
              column{even} = {c},
              column{3} = {c},
              column{5} = {c},
              column{7} = {c},
              column{9} = {c},
              column{11} = {c},
              cell{2}{1} = {c},
              cell{1}{3} = {c=3}{0.09\linewidth},
              cell{1}{7} = {c=3}{0.09\linewidth},
              cell{1}{11} = {r=2}{},
              cell{1}{12} = {r=2}{},
              hline{1,8} = {-}{},
              hline{2} = {1,3-5,7-9}{},
              hline{3} = {1,3-5,7-9,11-12}{},
            }
            \textbf{Total Exposure Time (s)} &  & $\mathbf{10^{-3}}$      &                 &                &  & $\mathbf{10^{-5}}$      &                 &                &  & \textbf{Avg. Run Time (s)} & \textbf{Sampling Ratio (\%)} \\
            \textbf{Metric}                  &  & PSNR$\uparrow$ & EMD$\downarrow$ & SSIM$\uparrow$ &  & PSNR$\uparrow$ & EMD$\downarrow$ & SSIM$\uparrow$ &  &                            &                              \\
            Push broom Sensor (PB)           &  & 26.7           & 2.0e-1          & 0.53           &  & 6.94           & 3.1e-1          & 0.029          &  & \textbf{0}                 & 100                          \\
            PB + HyperColorization           &  & 40.6           & 6.55e-3         & \textbf{0.961} &  & 33.2           & 1.0e-2          & 0.879          &  & 3.73                       & 10                           \\
            Cao et al.                       &  & 39.7           & 8.0e-3          & 0.934          &  & 31.8           & 1.5e-2          & 0.823          &  & 13.91                      & 3                            \\
            SASSI                            &  & 38.1           & 1.1e-2          & 0.940          &  & 34.4           & 1.4e-2          & 0.848          &  & \textbf{0.51}              & 3                            \\
            HyperColorization                &  & \textbf{41.0}  & \textbf{6.2e-3} & 0.955          &  & \textbf{35.9}  & \textbf{8.7e-3} & \textbf{0.946} &  & 3.97                       & 3                            
            \end{tblr}
            }
            }
            \caption*{HyperColorization boosts performance of traditional pushbroom sensors by allowing targeted exposure of color samples, exceeding traditional spatial-spectral resolution tradeoffs by allocating a short time budget to a smaller number of pixels before recovering spatial detail with a grayscale guide image. HyperColorization also outperforms the spectral sample propagation techniques introduced in hybrid approaches SASSI \cite{saragadam2021sassi} and Cao et al. \cite{cao2011high}, with HyperColorization results reported here under uniform whisk broom sampling.}
            \label{tab:comparisontable}
        \end{table}

\begin{figure}[ht!]
            \centering\includegraphics[width=12cm]{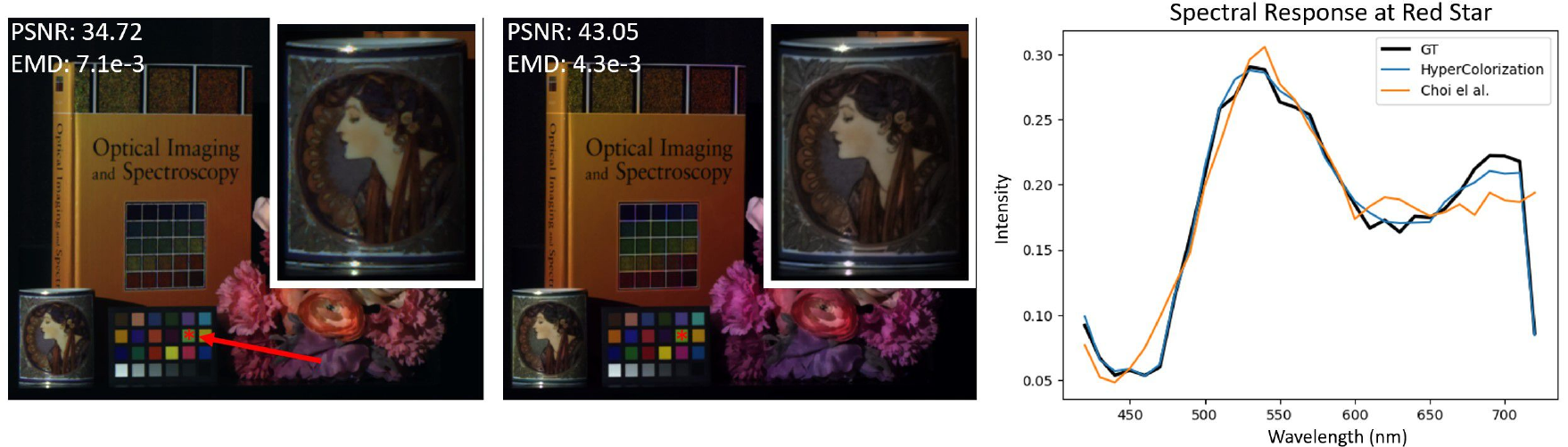}
            \caption{Comparison of Choi et al. \cite{choi2017high} and HyperColorization (3\%  guided whisk broom sampling, no noise added). HyperColorization reconstructs higher quality HSIs with dramatically less wall time (on the order of seconds vs. hours), but requires a more complicated optical system that involves two cameras.}
            \label{fig:compcassi}
        \end{figure}

        \FloatBarrier
        Another important class of spectral imagers is snapshot cameras. We benchmarked our technique against Choi et al. \cite{choi2017high}, who used a coded aperture snapshot spectral imager  system. With Choi et al.'s approach the spectral image is reconstructed using a single image which was captured through a dispersing prism and a coded aperture. The inverse problem solved by such cameras is highly ill-posed, however, they simplify the optical system and the data collection process. With Choi et al., due to the difficulty of the inversion, the reconstruction takes around 12 hours, which is significantly longer than the hybrid camera systems previously discussed. Fig. \ref{fig:compcassi} shows an example comparison of HyperColorization and Choi et al., demonstrating our sharper recovery of edges and texture while also obtaining more faithful spectral curves. 
\FloatBarrier
\section{Conclusion}
We presented a novel hyperspectral imaging framework called HyperColorization that enables traditional hyperspectral cameras such as whisk and push broom scanners to break the time-space-wavelength resolution trade-off through ‘colorizing’ a grayscale image by using spatially sparse spectral clues. We demonstrated that colorization within a learned subspace yields superior results in the presence of noise, and the optimal dimensionality of this subspace can be estimated directly from noisy spectral measurements. Grayscale-image-guided sampling algorithms offer slight improvements over uniform sampling patterns. Finally, our analysis of the trade-off between sampling ratio and exposure time per pixel suggest that prioritizing higher sampling ratios is more advantageous, as our framework provides effective tools for managing noise but lacks the ability to recover under-sampled images.



Next steps in improving HyperColorization involve refining grayscale-image-guided sampling algorithms, integrating spatial basis models into the colorization process, enhancing algorithm robustness for varied sampling scenarios, and devising heuristics to determine optimal sample numbers. Another interesting venue is to venture beyond the visible spectrum, particularly into the NIR and short-wave-infrared (SWIR) range. In these bands, we are not bound by the lower dimensional subspace seen in the visible range, creating more challenges but also greater rewards. In these spectral bands, there is a potential to develop spectral densification algorithms that support imaging modalities capable of recovering from compressed or sparse representations along the spectral dimension, such as the ones discussed in \cite{zhang2021deeply, feng2023superposition}. This would potentially decrease the necessary amount of measurements they need to take to get accurate reconstructions.

Computational imaging systems have enabled the development of snapshot hyperspectral imagers. Despite their low data collection times, the computational decoding of the captured data can still be time-consuming, posing practical limitations. While our framework has been primarily discussed in the context of whisk and push broom scanners, a key advantage of our approach lies in its adaptability to other imaging modalities as a post-processing step. In our future work, we aim to build upon this framework to decrease the computational cost and mathematical complexity associated with the decoding process in computational imagers. We are hoping to achieve this by potentially alleviating the decoding step’s ill-posed nature by isolating the measurements to enable the division of the intricate inversion task into more manageable sub-problems.




\begin{backmatter}
    \bmsection{Funding} This project was supported by a gift from Dolby.

    \bmsection{Acknowledgments}
    We would like to thank Yi-Chun Hung, Okyanus Oral, Vishwanath Saragadam, Regina Eckert, Jon McElvain for their valuable time and input.
    \bmsection{Disclosures}
    The authors declare no conflicts of interest.
    \bmsection{Data Availability Statement}
    Data underlying the results presented in this paper are available in Ref. \cite{github}.
    \bmsection{Supplemental document}
    See Supplement 1 for supporting content.
\end{backmatter}

\bibliography{sample}

\end{document}